\documentclass[11pt,a4paper]{article}
\usepackage[hyperref]{eacl2021}
\usepackage{times}
\usepackage{latexsym}

\usepackage{graphicx}
\usepackage{mathtools}
\usepackage{multirow}
\usepackage{bm}
\usepackage{hyperref}       % hyperlinks
\usepackage{url}            % simple URL typesetting
\usepackage{booktabs}       % professional-quality tables
\usepackage{amsfonts}       % blackboard math symbols
\usepackage{nicefrac}       % compact symbols for 1/2, etc.
\usepackage{microtype}      % microtypography

\DeclareMathOperator*{\argmax}{argmax}

\usepackage[bottom]{footmisc}
\usepackage{titlesec}
\usepackage{microtype}

\aclfinalcopy % Uncomment this line for the final submission
\title{Discovering Useful Sentence Representations from Large Pretrained Language Models}

\author{Nishant Subramani\\
  Scale AI\\
  \texttt{nishant.subramani@scale.com} \\\And
  Nivedita Suresh\\
  Arrive\\
  \texttt{nive@arriveorigin.com} \\}

\date{}

\begin{document}
\frenchspacing
\maketitle
\begin{abstract}
Despite the extensive success of pretrained language models as encoders for building NLP systems, they haven't seen prominence as decoders for sequence generation tasks.
We explore the question of whether these models can be adapted to be used as universal decoders.
To be considered "universal," a decoder must have an implicit representation for any target sentence $s$, such that it can recover that sentence exactly when conditioned on its representation.
For large transformer-based language models trained on vast amounts of English text, we investigate whether such representations can be easily discovered using standard optimization methods.
We present and compare three representation injection techniques for transformer-based models and three accompanying methods which map sentences to and from this representation space.
Experiments show that not only do representations exist for sentences from a variety of genres.
More importantly, without needing complex optimization algorithms, our methods recover these sentences \textit{almost perfectly without fine-tuning the underlying language model at all}.
\end{abstract}

\section{Introduction}
Recently, pretrained language models such as ELMo, BERT, and T5 have seen widespread success as encoders for a variety of natural language processing tasks often with little or no finetuning~\cite{peters2018deep, devlin2019bert, raffel2019exploring}. 
However, this has not transferred to decoders, i.e. most decoders for sequence generation tasks are task-specific and are trained from scratch~\cite{Nallapati2016AbstractiveTS, johnson-etal-2017-googles, aharoni-etal-2019-massively}.
We explore whether pretrained language models can be modified to be used as "universal" decoders.

For a decoder to be considered "universal", it must be able to successfully recover a sentence when conditioned on its implicit sentence representation.
Such a decoder would provide many benefits: make training text generation models on little amounts of annotated data possible, allow considerable parameter sharing in memory- and data-limited environments, and improve zero-shot text generation performance.
Imagine you are tasked with building a Kurdish to English translation model.
You find that there's very little parallel data on this language pair to learn from and realize that an end-to-end trainable sequence-to-sequence model cannot be fit well.
If you had a universal decoder, you may be able to train a Kurdish encoder, which is much smaller than the entire sequence-to-sequence model, and optimize it to work with the universal decoder.

In this work, we take an initial step towards evaluating whether large pretrained language models can be used as universal decoders without finetuning.
We first define the \textit{sentence space} of a transformer language model, GPT-2~\cite{radford2019language}, and reparametrize each point in this space to a lower-dimensional point by adding a single bias term $\boldsymbol{z}$ to various locations in the model.
Keeping the language model fixed, we optimize $\boldsymbol{z}$ to maximize the likelihood of the original sentence $\boldsymbol{x}$ and recover $\boldsymbol{x}$ from $\boldsymbol{z}$ in order to evaluate how useful the representation is.
In other words, we \emph{reverse-engineer} a sentence representation that generates the target sentence.

\begin{figure*}[t]
\includegraphics[width=\textwidth]{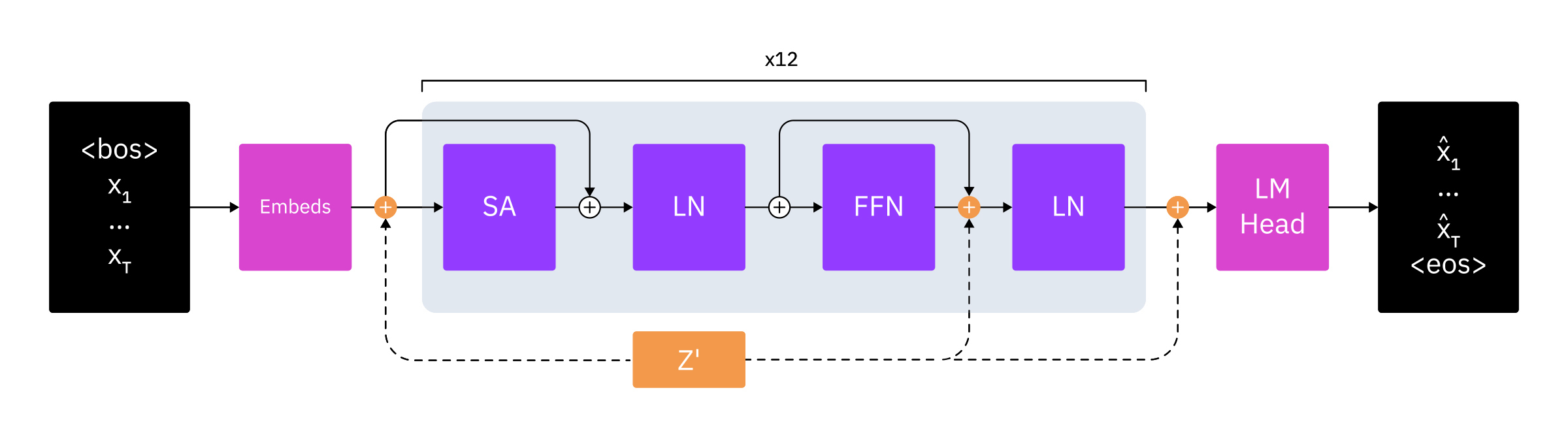}
\caption{We add a bias $Z'$ based on Equation~\ref{eq:rep-injection-mech} to three different locations in GPT-2: to the embedding, to the transformer layers, and before the language modeling head. Here 'Embeds' refers to the embedding, 'SA' to self-attention, 'LN' to layer normalization~\citep{Ba2016LayerN}, 'FFN' to a fully-connected layer, and 'LM Head' to the last fully-connected layer.}
\label{fig:process}
\end{figure*}
Our experiments uncover that we can achieve nearly perfect recoverability with a reparametrized sentence space of dimension equal to the latent dimension of the language model.
That is to say, for nearly all sentences, there exists at least one relatively low-dimensional vector that, by itself, can recover the sentence of interest nearly exactly.
Further, we show that this holds for text from a variety of genres ranging from books to news to movie quotes to Wikipedia.
We learn that discovering nearly perfect representations is relatively easy using simple optimization with Adam~\citep{kingma2014adam}, unlike previous work~\citep{subramani2019can}.
Our experiments show that recoverability increases as the dimensionality of the reparametrized space increases and decreases with increased sentence length, i.e. recoverability is lower for longer sentences.
Using PCA, we find that the reparametrized sentence space does not lie on a lower-dimensional linear manifold, and confirms that the intrinsic dimension of the reparametrized space is approximately equal to the latent dimension of the language model.

\section{Learning Sentence Representations}
Below, we discuss background on transformer-based language models and characterize how these models represent sentences~\citep{vaswani2017attention}.
We show how to reparametrize this space into a lower-dimensional space and define the notion of the recoverability of a sentence in this reparametrized space.
We show these for GPT-2, but indicate how our methodology is model-agnostic.

Transformer language models such as GPT-2, represent a sentence $\boldsymbol{x} = x_1, \ldots, x_T$ as a sequence of hidden states $\boldsymbol{h_1}, \ldots, \boldsymbol{h_{T}}$, which come from the final layer of the transformer model.
Since $\boldsymbol{h_i} \in \mathbb{R}^d$, where $d$ is the latent dimension of the language model, the model encodes $x_1, \ldots, x_T$ in a sentence space $\mathcal{H} \in \mathbb{R}^{d\times T}$.
Representations in this sentence space are sequence length dependent, making comparisons between sentences with differing lengths inequitable and measuring the efficacy of using an unconditional language model as a universal decoder impossible.
To resolve these issues and to make analysis easier, we reparametrize the sentence space into a lower-dimensional and sentence-length agnostic vector space.

\subsection{Representation Space}

We propose to reparametrize the original sentence space $\mathcal{H} \in \mathbb{R}^{d\times T}$ to $\mathcal{Z} \in \mathbb{R}^{d'}$, mapping a sentence length dependent, high-dimensional vector space into a lower dimensional, sentence-length agnostic vector space of dimension $d'$.
In our experiments, $d' \leq d$.
We do this by adding a bias term $\boldsymbol{z} \in \mathbb{R}^{d'}$ to the fixed language model and find a $\boldsymbol{\hat{z}}$ that minimizes the cross entropy loss of the sentence.
We inject $\boldsymbol{z}$ by using a projection matrix $W_z \in \mathbb{R}^{d \times d'}$, which is never trained and is fixed throughout.
\begin{equation}\label{eq:w-z}
    W_z = [I_{d'}; W_{mix}]^\top
\end{equation}
Here, $W_{mix} \in \mathbb{R}^{d' \times (d-d')}$ is a probability weight matrix where the columns sum to 1, where we sample each entry from a standard Gaussian and compute a softmax over columns.
We randomly permute the independent and dependent components of $W_z$ to avoid an arbitrary, fixed ordering of columns.

Our reparametrization must give us the ability to project a sequence of tokens $\textbf{x} = x_1, \ldots, x_T$ into a representation $\boldsymbol{z}$ (sentence encoding) and to recover $\textbf{x}$ from $\boldsymbol{z}$ (sentence recovery) via the language model.
Without this property, we cannot measure recoverability.
Imagine a task-specific encoder trained to produce context for a conditional generation task.
The output of such an encoder resembles the $\boldsymbol{z} \in \mathcal{Z}$ we wish to discover.
With our reparameterization approach, we expect $\boldsymbol{z}$ to encode the target sentence using sentence encoding and regenerate it using sentence recovery.

\subsection{Representation Injection}

We experiment with three $\boldsymbol{z}$ injection locations: embedding (embed), each layer of the transformer (layers), and language model head (head).
See Figure~\ref{fig:process} for details.
We also experiment with three representation injection mechanisms that transform $\boldsymbol{z}$ to $\boldsymbol{z'}$ and inject $\boldsymbol{z'}$ into the language model: no ensembling, attention-based ensembling, and interleaved ensembling.
Ensembling splits up $\boldsymbol{z}$ into $k$ experts and allows those $k$ experts to work together to learn a sentence representation.
Here, $\boldsymbol{z}$ is split up into a matrix $Z \in \mathbb{R}^{\frac{d'}{k} \times k}$ and $W_z \in \mathbb{R}^{d \times \frac{d'}{k}}$.
In no ensembling, $k = 1$, so $Z = \boldsymbol{z}$.
In attention-based ensembling, we use soft-attention with the previous layer's hidden state~\citep{DBLP:journals/corr/BahdanauCB14}, allowing the model to learn an adaptive combination of the $k$ vectors per input token.
In interleaved ensembling, we use the first vector for the first token, the second for the second token, until we reach $k$. 
After we process the $k^{\text{th}}$ token, we start the process over again with the first vector.
This way, each of the $k$ vectors are responsible for only every $k^{\text{th}}$ token.
To do this, we use $W_{\text{int}} \in \mathbb{R}^{T \times k}$, which comprises of $\frac{T}{k}$ many $I_k$ matrices concatenated together and the first $T$ rows chosen.
Below are the equations for no ensembling, attention-based ensembling, and interleaved ensembling respectively:
\begin{align}
\label{eq:rep-injection-mech}
    &Z' = \begin{cases} 
        W_z Z, \\
        \emph{softmax}(H_{t-1}(W_z Z))(W_z Z)^\top,\\
        W_{int}(W_z Z)^\top,\\
   \end{cases}
\end{align}

\subsection{Sentence Encoding \& Recovery}
In sentence encoding, we project a sentence $\boldsymbol{x}$ into a representation $\boldsymbol{z}$ via the language model $\Theta_{LM}$ using Equation~\ref{eq:rep-injection-mech}.
We estimate $\boldsymbol{z}$ by maximizing the log probability of $\boldsymbol{x}$, while keeping $\Theta_{LM}$ fixed:
\begin{equation}
\label{eq:sentence-projection}
    \hat{\boldsymbol{z}} = \argmax_{\boldsymbol{z} \in \mathcal{Z}} \sum_{t=1}^T \log p(x_t | \boldsymbol{x}_{<t}, \boldsymbol{z})
\end{equation}
Here, we represent the entire sentence $\boldsymbol{x}$ with a single $\boldsymbol{z}$. 
Since this objective function is highly non-convex and could potentially lead to many local optima, we randomly initialize $\boldsymbol{z}$, $n$ times and measure recoverability over them.
Our experiments reveal that different $\boldsymbol{z}$'s can recover the original sentence perfectly, although recoverability is somewhat sensitive to initialization.

Sentence recovery aims to recover the original sentence $\boldsymbol{x}$ from $\boldsymbol{z} \in \mathcal{Z}$.
In essence, we find the most probable sentence $\textbf{x}$ under the model, $\Theta_{LM}$.
Our experiments show that beam search and greedy decoding perform similarly even with different beam widths.
Therefore, all results presented here use greedy decoding without assuming a true length.
We stop when decoding produces either an end-of-sentence token or 150 consecutive tokens.

\section{Measuring the Effectiveness of Sentence Representations}
We want our sentence representations to be unique and implicit for each target sentence $s$ such that when our language model is conditioned by our representation, it can recover $s$ exactly.
Our formulation does not require a bijective mapping, only a surjective mapping between the sentence representation $\boldsymbol{z}$ and the original sentence $s$.
We measure the effectiveness of these representations through the lens of recoverability using three common metrics~\cite{subramani2019can}.

\subsection{Recoverability Metrics}
When measuring recoverability, we estimate how much information our representation $\boldsymbol{z}$ retains about the target sentence $s$.
To estimate how much relevant information about generation our representations contain, we measure token-level exact match, prefix match, and Smoothed BLEU using the target sentence $s$ and our reconstruction of it, $\hat{s}$~\cite{subramani2019can}.
Token-level exact match calculates the average number of correct tokens in a candidate sentence.
Prefix match measures the longest consecutive sequence of tokens from the beginning of the sentence which are recovered correctly as a proportion of the length of the target sentence.
This is relevant because auto-regressive natural language generation has a very strong left-to-right tendency due to decoding occurring left-to-right for English and other left-to-right languages~\citep{subramani2019can}.
Smoothed BLEU provides a smoother approximation to token-level exact match and is a popular metric in evaluating conditional language modeling tasks such as machine translation~\citep{papineni2002bleu, chen2014systematic}.
To measure smoothed BLEU, we use sacrebleu's exponential smoothing with the WMT standard 13a tokenization~\citep{post2018call}.
We use $n$ random initializations and recover the same target sentence $\boldsymbol{x}$ from each of them, computing mean scores to measure initialization variability.
In addition, we evaluate the maximum scores from those $n$ random initializations across our metrics: \textbf{EM-Max}, \textbf{PM-Max}, and \textbf{BLEU-Max}.

\subsection{Analyzing Intrinsic Dimension}
Under the lens of recoverability, we define the intrinsic dimension of the reparametrized sentence space to be the smallest dimension of $\boldsymbol{z}$ ($d'$) that produces a specific target recoverability $\tau$~\cite{bojanowski2018optimizing, subramani2019can}:

\begin{equation}\label{eq:intrinsic-dim}
    \hat{d}'(\theta,\tau) = \min_{d'}\left\{ d':\overline{\emph{BLEU}}(D|(d',\theta))>\tau\right\}
\end{equation}
Here, $\overline{\text{BLEU}}$ is the target recoverability measure for dimension $d'$ for model $\theta$ and is computed as:
\begin{align}\label{eq:recoverability-measure}
    \overline{\text{BLEU}}(D_x | \theta, d') &= \frac{\sum_{\boldsymbol{x} \in D_x}\sum_{i=0}^{n}BLEU(\boldsymbol{\hat{x}_i}, \boldsymbol{x})}{|D_x|\cdot n}\\
    \overline{\text{BLEU}}(D | \theta, d') &= \frac{1}{|D|} \sum_{D_x \in D} \emph{BLEU}(D_x|\theta, d')
\end{align}
Here, $|D|$ is the number of corpora, $|D_x|$ is the number of sentences in each corpus, $n$ is the number of different random initializations of $\boldsymbol{z}$ per sentence per corpus, and $\boldsymbol{\hat{x}}$ is the predicted sentence.

In addition, we analyze the intrinsic dimensionality of $\mathcal{Z}$ using principal component analysis by transforming $\mathcal{Z} \in \mathbb{R}^{d'}$ into orthogonal basis vectors.
Equipped with these orthogonal bases, we can measure how many components are required to capture a proportion $p$ of the variability in the data using cumulative explained variance.

\begin{table*}[t!]\small\centering
\begin{tabular}{@{}l|lll|llllll@{}}
\toprule
                                            & Init           & Location     & Ensembling    & EM & PM & BLEU & EM-max & PM-max & BLEU-max\\ \midrule
{\multirow{2}{*}{I}}                        & L2             & All          & None          & 98.1 & 98.4 & 98.1 & 100.0  & 100.0  & 100.0\\
                                            & \textbf{Xavier}& \textbf{All} & \textbf{None} & \textbf{99.0} & \textbf{99.0} & \textbf{98.9} & \textbf{100.0}  & \textbf{100.0}  & \textbf{100.0}\\ \midrule
\multirow{4}{*}{II}                         & Xavier         & Embed     & None         & 44.8 & 44.9 & 44.6 & 72.3   & 72.2   & 71.9\\
                                            & Xavier         & +Layers   & None         & 98.8 & 98.8 & 98.8 & 100.0  & 100.0  & 100.0\\
                                            & Xavier         & Head      & None         & 4.1  & 3.8  & 3.3  & 4.1    & 3.8    & 3.3\\
                                            & \textbf{Xavier}& \textbf{All} & \textbf{None} & \textbf{99.0} & \textbf{99.0} & \textbf{98.9} & \textbf{100.0}  & \textbf{100.0}  & \textbf{100.0}\\ \midrule
\multirow{3}{*}{III}                        & Xavier         & All       & Attention (k=2)    & 82.8 & 82.2 & 83.0 & 97.3 & 97.3 & 97.3\\
                                            & Xavier         & All       & Attention (k=4)    & 49.4 & 49.0 & 49.5 & 79.2 & 79.0 & 79.9\\
                                            & Xavier         & All       & Interleave (k=2)   & 69.3 & 68.0 & 69.7 & 82.2 & 81.3 & 82.6\\
                                            & Xavier         & All       & Interleave (k=4)   & 65.4 & 65.0 & 65.4 & 89.2 & 89.1 & 89.2\\
                                            & \textbf{Xavier}& \textbf{All} & \textbf{None} & \textbf{99.0} & \textbf{99.0} & \textbf{98.9} & \textbf{100.0}  & \textbf{100.0}  & \textbf{100.0}\\ \bottomrule
\end{tabular}
\caption{Recoverability results for Phase I on SRC}
\label{tbl:SRC}
\end{table*}

\section{Experimental Setup}
\paragraph{Data Collection}
For experiments on sentence recoverability, we create a dataset which combines four corpora from different genres: movie dialogs (movies), classic books (books), news articles (news), and Wikipedia (wiki).
For movies, we choose the Cornell Movie Dialogs corpus~\citep{DanescuNiculescuMizil2011ChameleonsII}, which consists of fictional conversations from 617 raw movie scripts.
We choose NLTK's Gutenberg dataset for our books portion, which consists of a subset of texts from Project Gutenberg~\citep{lebert2008project}.
Our news subset comes from the Gigaword dataset for abstractive summarization~\citep{graff2003english}, consisting of 3.8 million articles.
Lastly, our Wikipedia portion comes from WikiText-103~\citep{Merity2017PointerSM}, a dataset with 28,475 verified articles. 
For movies, news, and wiki, we extract sentences from its pre-specified validation set.
For books, since NLTK's Gutenberg dataset lacks a pre-specified data split, we consider the entire dataset.

\paragraph{Data Preprocessing}
We sentence tokenize all of our datasets using NLTK's sentence tokenizer.
Next, we randomly sample 16 sentences from each corpus, making sure sentences are between 5 and 100 words according to NLTK's word-level, regular expression tokenizer.
We call this the small recovery corpus (SRC).
To construct a larger corpus, the large recovery corpus (LRC), we group sentences by sentence length into 8 bins: 5-10, 10-15, 15-20, 20-25, 25-30, 30-35, 35-40, and 40-100, and randomly sample 64 sentences from each of the bins, ensuring that no sentences overlap between LRC and SRC.
Lastly, we create a third corpus that we call the gibberish recovery corpus (GRC), by sampling tokens uniformly at random with replacement from the GPT2 vocabulary such that we have 8 gibberish sentences in each of the 8 sentence length bins above similarly to~\citet{subramani2019can}.

\paragraph{Phase I: Experimental Phase}
We use SRC to evaluate the best initialization technique (I), injection location (II), and ensembling strategy (III) in an iterative manner in this order.
Refer to Table~\ref{tbl:SRC} for details.
In these experiments, we use stochastic gradient descent with Adam with a learning rate of 0.01~\citep{kingma2014adam}, maximum number of optimization steps of 1000, learning rate decay with a plateau with a patience of 3 and decay factor of 0.8, dimensionality of $\boldsymbol{z}$ of 768, and $n$, the number of random $\boldsymbol{z}$ initializations, of 4.
Motivated by looking at a few iterations of sentence encoding, we stop optimization early if the learning rate decays to $1\mathrm{e}{-5}$.
We also stop optimization early if mean cross entropy loss reaches $\min(0.1, \frac{2}{T})$, where $T$ is sequence length.
This heuristic is not crucial, but allows experimentation to run quickly without a degradation in performance.

\paragraph{Phase II: Testing Phase}
We use LRC to evaluate recoverability in order to estimate the intrinsic dimension of $\mathcal{Z}$ (IV).
Using the same hyperparameters from phase I and choosing the best initialization method, injection location, and ensembling strategy, we estimate the intrinsic dimension of the reparameterized sentence space by varying the dimension of $\boldsymbol{z}$, $d'$, to be 192, 384, 576, and 768.

\section{Results \& Analysis}
\paragraph{Recoverability on SRC}
Experiment I indicates that initialization strategy does not affect performance significantly, but xavier normal performs better than l2 normalization.
Injection location, on the other hand, has a tremendous effect on performance.
Injecting $\boldsymbol{z}$ at the language modeling head alone leads to poor performance as the final fully connected layer is severely bottlenecked in terms of capacity~\citep{Yang2017BreakingTS}, but injection into the embedding alone allows the transformer model to work with $z$ and learn from it --- leading to a 10x improvement over just the lm head.
Above all of this, injecting into the transformer model at every layer including the embedding virtually solves the task, achieving nearly perfect recoverability across the board.
We theorize that this is due to the model continuously seeing $\boldsymbol{z}$ at each layer, which make optimization easier and more stable.
We find that additionally injecting into the head leads to a slight increase in recovery, so we inject $\boldsymbol{z}$ at all three places for all of the following experiments.

Representation injection mechanisms also have a large impact on recovery: both attention-based and interleaved experts perform significantly worse than no experts.
These methods suffer from the fact that splitting $\boldsymbol{z}$ into $k$ smaller vectors reduces capacity and makes retaining information more difficult.
See Table~\ref{tbl:SRC} for details.
We find that regardless of experimental criteria, all six metrics are extremely consistent and correlate nearly perfectly to one another.
As a result, we only report BLEU score means for the remainder of experiments.

\begin{figure}[h!]
\centering
\includegraphics[width=\linewidth]{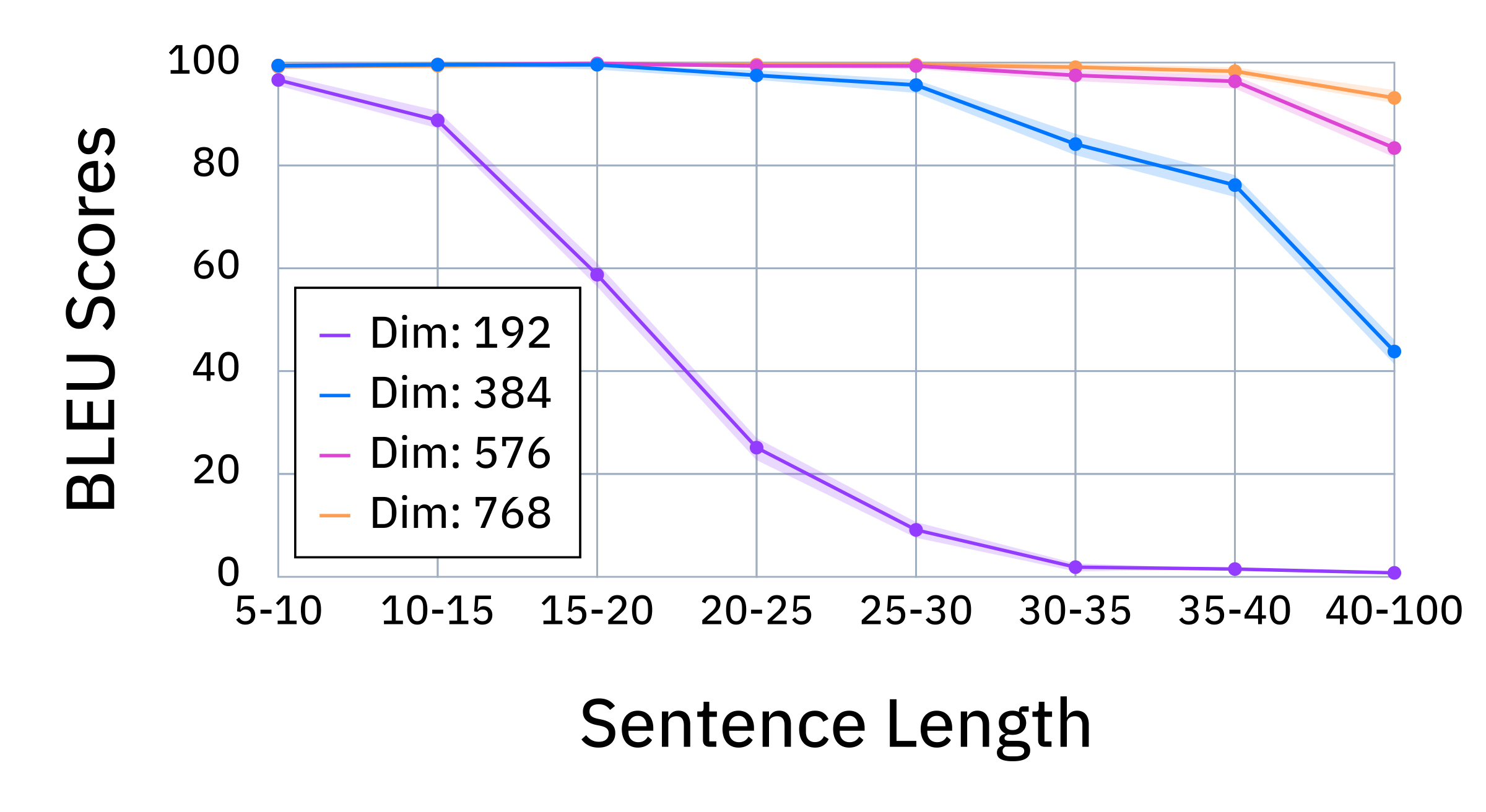}
\caption{Plot of sentence length vs. BLEU score on LRC for experiment IV with error regions of $\pm{\sigma}$.}
\label{bleu-intrinsic-dim}
\end{figure}

\paragraph{Intrinsic Dimension via Recoverability:}
In experiment IV, we estimate the intrinsic dimension of $\mathcal{Z}$.
We observe that $\overline{\emph{BLEU}}$ increases as $d'$ increases until $d' = 768$, where $\overline{\emph{BLEU}}$ is nearly perfect --- hinting that the intrinsic dimension of $\mathcal{Z}$ is approximately 768.
However, a lower-dimensional representation can recover most sentences, dropping off as sentence length increases, see Figure~\ref{bleu-intrinsic-dim}.
This is well-known; the number of bits needed to encode a sequence grows linearly with its length.
We observe low variances in our estimations, especially as $d'$ increases, indicating that the differences in $\overline{\emph{BLEU}}$ for different values of $d'$ are statistically significant.

\begin{figure}[h]
\includegraphics[width=\linewidth]{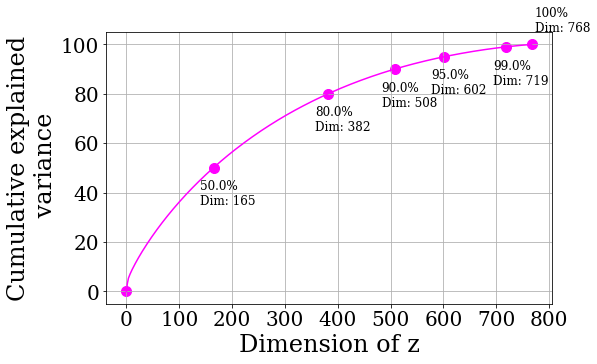}
\caption{Cumulative explained variance plot under PCA with on LRC with number of components equal to $d'=768$.}
\label{fig:bleu-pca-768}
\end{figure}

\paragraph{Intrinsic Dimension via PCA:}
We pick the best performing $\boldsymbol{z}$ under \textbf{BLEU-max} for each sentence from experiment IV with $d'=768$ and apply PCA to retain 768 components ($n_{comp}$).
We observe that both intrinsic dimension experiments via PCA and via recoverability show similar patterns.
The shape of the curve in Figure~\ref{fig:bleu-pca-768}, hints that $\mathcal{Z}$ does not lie on a lower-dimensional linear manifold and that its intrinsic dimensionality is approximately 768.
$n_{comp} \approx 600$ explains almost 95\% of the data's variance, which supports our observations from experiment IV that shows $d'=576$ achieving nearly perfect $\overline{\emph{BLEU}}$ (Figure~\ref{bleu-intrinsic-dim}).

\begin{figure}[h]
\centering
\includegraphics[width=0.75\columnwidth]{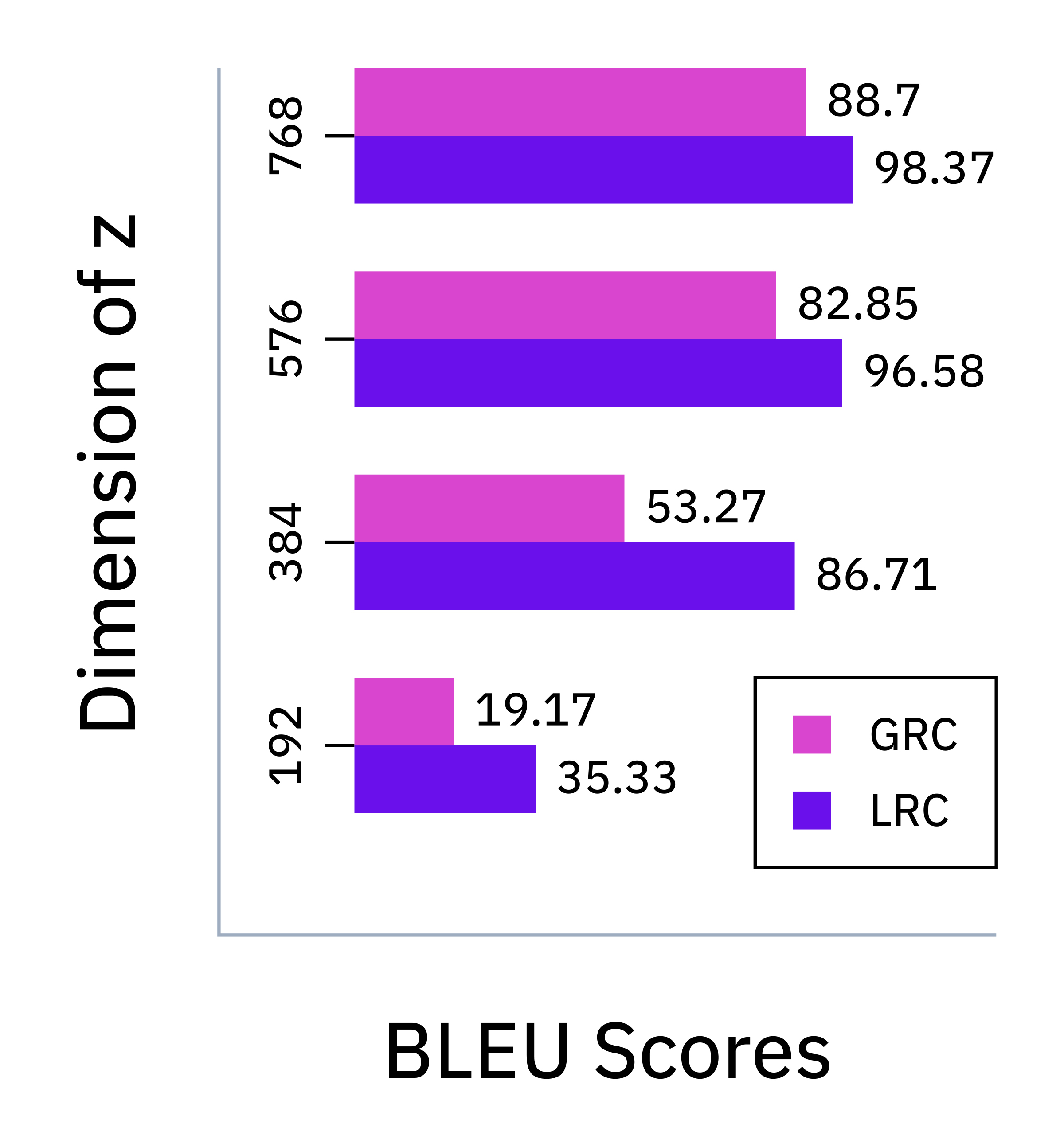}
\caption{$\overline{\emph{BLEU}}$ performance on LRC versus GRC for different dimensionalities of $\boldsymbol{z}$.}\label{fig:lrc-vs-grc-bleu}
\end{figure}

\paragraph{Recoverability on GRC:}
We run the intrinsic dimension experiment on the gibberish dataset (GRC) and find that performance on the real dataset exceeds that on the gibberish dataset for all dimensions.
This hints at the fact that although our representations memorize, they also leverage the language model.
Even though $\overline{\emph{BLEU}}$ for $d'=576$ and $d'=768$ for GRC seem high, the error on GRC is 5x that of LRC (Figure~\ref{fig:lrc-vs-grc-bleu}).

\begin{figure*}[t!]
\centering
\includegraphics[width=0.75\textwidth]{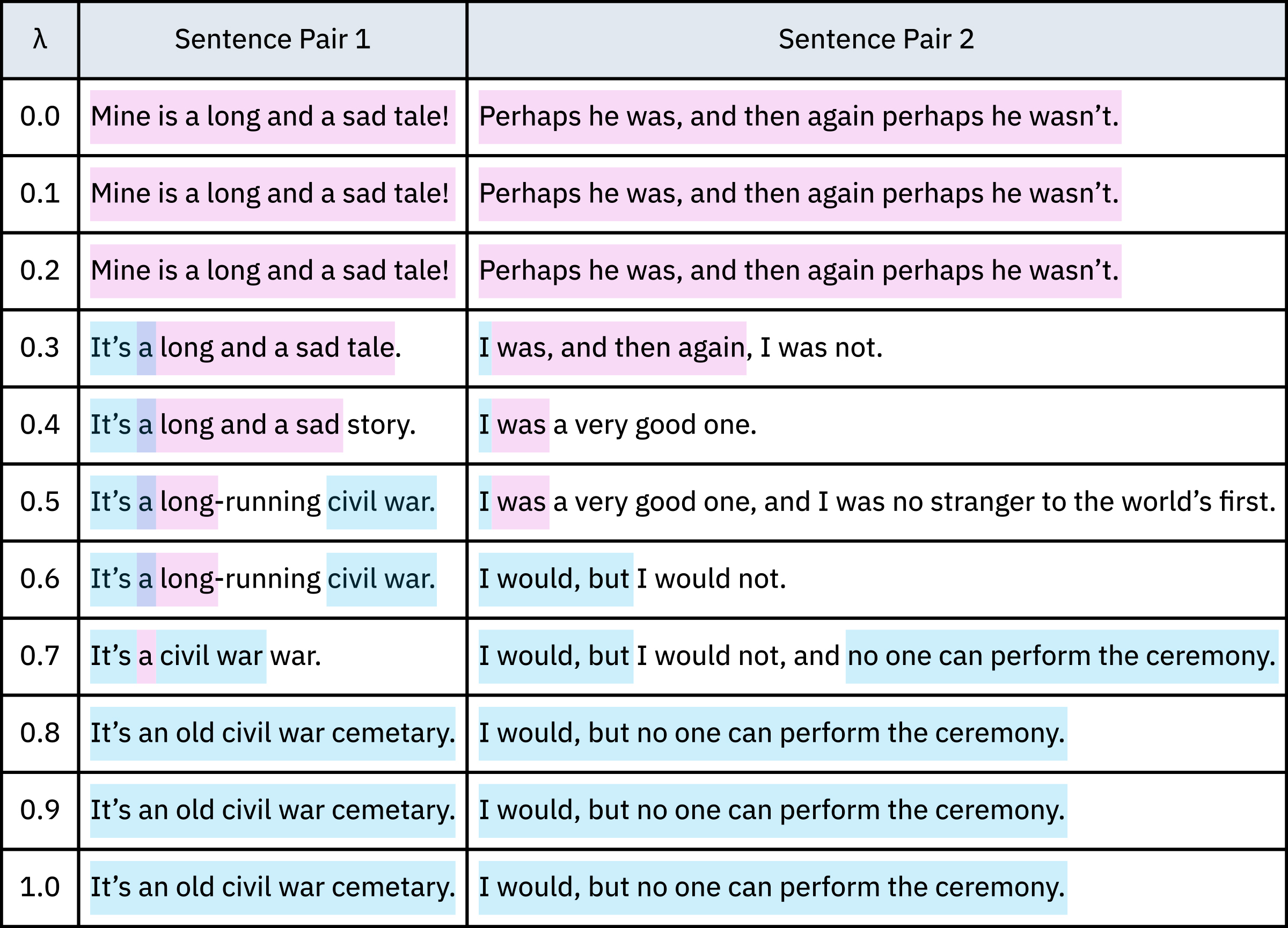}
\caption{Two linear interpolations between perfectly recovered pairs of representations. Pink indicates token overlap to the first sentence, while blue indicates token overlap to the second sentence.}
\label{fig:interpolation}
\end{figure*}

\paragraph{Interpolation:}
In Figure~\ref{fig:interpolation}, we show linear interpolations of two pairs of $\boldsymbol{z}$'s that recover sentences exactly.
The space is smooth with well-formed grammatical sentences occupying areas with $\lambda = [0.3, 0.6]$.
Our learned representations seem to have some synonym awareness: "tale" transforms to "story" in the first sentence pair and "long" transforms to "long-running" when referring to a war.
In the second sentence pair, we observe some notion of syntactical awareness: at the 0.7 mixture level the syntax of the first sentence is retained with mostly words from the second sentence.
Lastly, for each individual sentence there exists a $d$ dimensional volume that is fairly large.
This could indicate that nearly all sentences have some representative volume from which, if any vector was sampled, sentence recovery could generate that sentence exactly.

\begin{figure}[h]
\centering
\includegraphics[width=0.75\columnwidth]{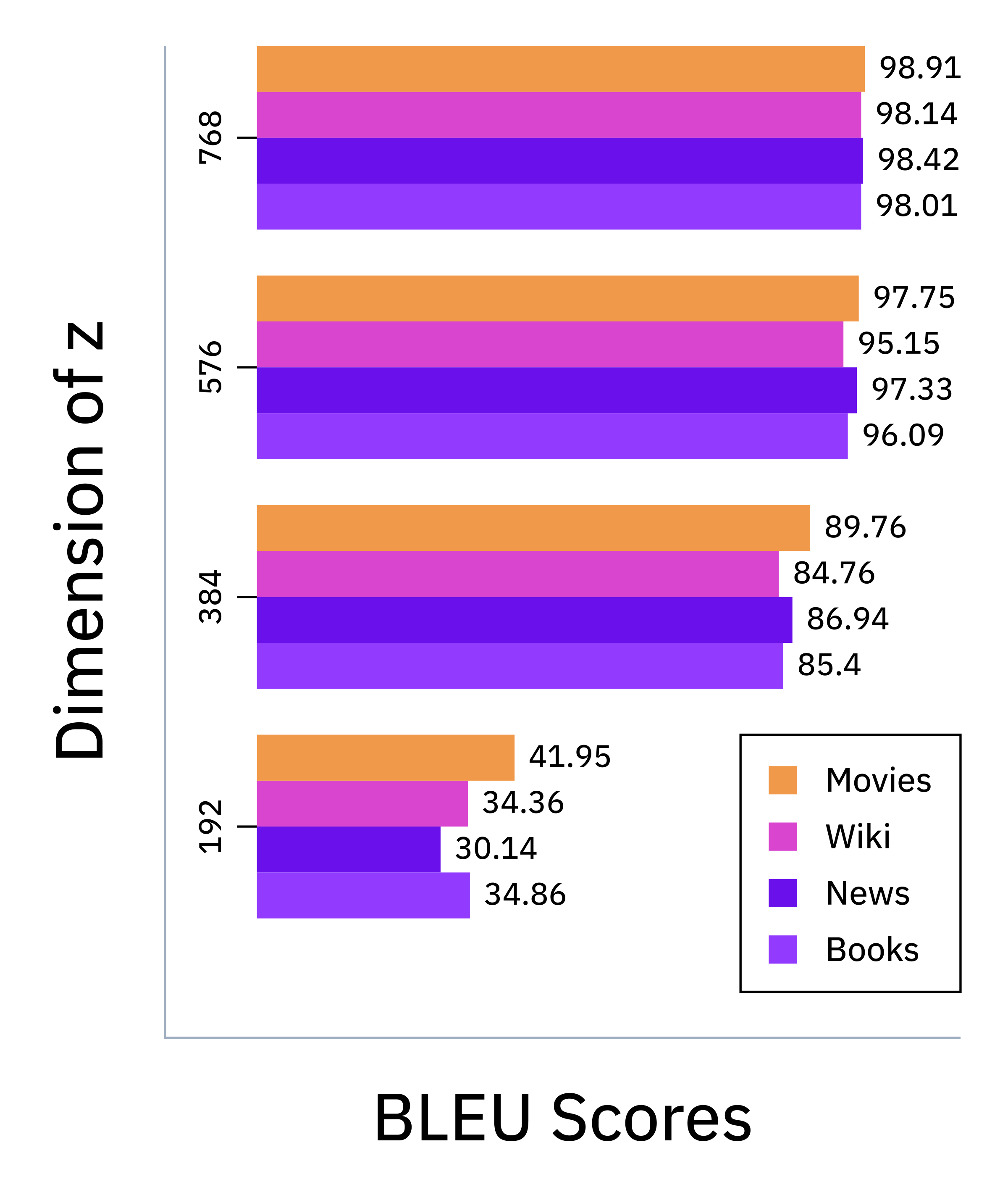}
\caption{$\overline{BLEU}$ performance on LRC stratified by genre for different dimensionalities of $\boldsymbol{z}$.}\label{fig:genre_bleu}
\end{figure}

\paragraph{Towards a Universal Decoder:}
We can discover representations, which exactly recover target sentences of interest in a low-dimensional space using Adam.
Other work found this impossible with $\overline{BLEU} < 1$ even for short sentences with less than 10 words, when applying an analogous technique on LSTM-based language models~\citep{subramani2019can}.
For sentences up to 100 words, we discover representations which achieve over 98 $\overline{BLEU}$, generalizing to text from a variety of genres (Figure~\ref{fig:genre_bleu}).
Our representations do not simply memorize, but actually leverage the fixed language model, leading to representations with some interpretability.
Lastly, interpolation experiments show that our reparametrized space has some synonym and syntactical awareness, while maintaining a strong prior for sentences to be mostly grammatically correct even in regions near the midpoint between two sentences.
As a result, our formulation and representation space analysis hints at the fact that unconditional language models have the potential to be used as universal decoders and that designing an encoder to learn these types of representations may be possible.

\section{Related Work}

\paragraph{General-purpose Decoders}
Large pretrained language models are used for extracting meaningful task-specific representations for different Natural language processing tasks. ~\citep{gulcehre2015using, Zoph2016TransferLF, Sriram2018ColdFT, nogueira2019passage}.
Other methods pretrain sequence-to-sequence decoders for tasks such as abstractive summarization and neural machine translation~\citep{edunov2019pre, song2019mass, chan2019kermit}.
None of these methods analyze sentence representations or evaluate the difficulty in discovering such representations.

\paragraph{Latent Space of Models}
Our notion of sentence space resembles work on generative latent optimization because we also perform inference on a implicit latent variable $z$, the sentence representation, using a fixed language model $\theta$~\citep{bojanowski2018optimizing}.
Using ideas about difficulty of latent variable optimization and interpolation from prior work on latent variable language models based on variational autoencoders~\citep{bowman2016generating}, denoising autoencoders~\citep{lewis2019bart}, generative adversarial networks~\citep{yu2017seqgan}, and plug-and-play models for image and text generation~\cite{nguyen2017plug, dathathri2019plug}, we develop our notion of the reparametrized sentence space $\mathcal{Z}$ and analyses that follow.
We focus on analyzing the sentence space of a fixed pretrained unconditional language model rather than training or fine-tuning.

\paragraph{Analysis of Language Models}
Many works focus on probing language models to understand what they know: evaluating their performance on question-answering or fill-in-the-blank tasks or evaluating how well they transfer these kinds of tasks~\citep{donahue2020enabling, tamkin2020investigating, hu2020systematic, gururangan-etal-2020-dont}.
We focus on understanding how these models represent sentences, the complexity of that representation, and how easily discoverable those representations are.
The goal of identifying complexity of a sentence representation resembles work that analyzes continuous bag-of-words representations with low-rank subspaces~\citep{mu-etal-2017-representing}.
\citet{Subramanian2018TowardsTG} learn latent representations based on general-purpose encoders for neural outlines and conclude that these outlines are informative for generation.
We focus on a different and more basic question, whether a pretrained language model has the potential to be used as a universal decoder.

Recently, there has been work on investigating whether LSTM-based language models have sentence representations from which they can recover the original sentence~\citep{subramani2019can}.
This work is the closest to ours.
We extend their work to transformer-based language models and improve upon their reparametrization leading to representations which are 5x smaller that still achieve nearly perfect recovery across a much greater variety of genres.
Furthermore, we show that our representations are easily discoverable using simple optimization rather than needing to use specialized conjugate gradient methods.

\section{Conclusion}
To evaluate whether unconditional language models have the potential to be used as universal decoders without fine-tuning, we introduce a reparametrized sentence space $\mathcal{Z}$.
In this space, a sentence is represented as a low-dimensional vector $\boldsymbol{z}$, which we use to condition a language model, which is optimized to generate that sentence during decoding.
We present two methods, sentence encoding and sentence recovery, which allow us to map a sentence to and from $\mathcal{Z}$.
Using these procedures, we evaluate whether we can discover representations that recover a sentence nearly perfectly.
Further, we measure the intrinsic dimension of $\mathcal{Z}$ under the lenses of recoverability and PCA.

We observe that such representations are easily discoverable with simple stochastic optimization, unlike prior work, even while varying genres of text.
We find that recoverability increases with the dimension of the reparametrized sentence space, reaching nearly perfect performance when equal to the latent dimension of the model.
Experiment IV shows that sentence length and recoverability are inversely related.
Analysis using PCA indicates that $\mathcal{Z}$ does not lie on a lower-dimensional linear manifold and confirms that the intrinsic dimension of $\mathcal{Z}$ is close to the latent dimension $d$ of the language model.
Our estimates for intrinsic dimension are upper-bounds, while the associated recoverabilities are lower-bounds due to the non-convexity of the objective function, the stochasticity of the sentence encoding step, and the approximate nature of greedy decoding.

Our sentence representation formulation has many useful properties: nearly perfect recoverability, smoothness in the representation space, and easy representation recovery (simple optimization) --- indicating the potential for GPT-2 to be used as a universal decoder.
As a result, a next step could be to design an encoder which would learn mappings from its task-specific input representation space to our reparametrized sentence space.
Another avenue for future work could be adapting this approach to work on more transformer-based language models.

Having a universal decoder could result in tremendous progress for low-resource sequence generation tasks from both a data and memory perspective.
Translation tasks such as Kurdish to English are an ideal use case because they have little parallel data, but have a target language (English) with abundant monolingual data.
Our reparametrized sentence space formulation and the potential of using an unconditional language model as a universal decoder may drive progress in building more generalizable systems with large-scale language models.
These models may encode and amplify some unwanted biases present in both the data sources and the organizations building them.
Many language models are used in commercial NLP applications without much concern for bias mitigation, but our approach could be modified to attempt to mitigate some of these biases.
As with sequence generation models broadly, there are always significant risks of this research aiding misinformation spread.
Our work indicates that well-trained large language models have a sentence representation for any well-formed target sentence, so malicious attackers could build harmful sequence generation systems in news headline summarization and dialog to name a few.

\section*{Acknowledgments}
We gratefully acknowledge Ty Wilkins for many of the visualizations and plots in this paper.
We thank members of the Scale AI machine learning team and members of the Arrive engineering team for feedback on iterations of this work.

\bibliography{anthology,eacl2021,references}
\bibliographystyle{acl_natbib}

\clearpage
\begin{table*}[t!]\small\centering
\begin{tabular}{@{}l|l|llllll@{}}
\toprule

Dataset & Dimension & EM & PM & BLEU & EM-max & PM-max & BLEU-max \\ \midrule
{\multirow{2}{*}{Complete}} 
& 192 & 35.10 & 34.71 & 35.33 & 45.11 & 44.25 & 45.12 \\
& 384 & 86.33 & 86.20 & 86.71 & 93.90 & 93.81 & 94.25 \\
& 576 & 96.19 & 96.10 & 96.58 & 98.50 & 98.44 & 98.87 \\
& 768 & 97.99 & 97.96 & 98.37 & 99.32 & 99.32 & 99.68 \\ \midrule

{\multirow{2}{*}{Books}} 
& 192 & 34.77 & 34.25 & 34.86 & 44.92 & 43.88 & 44.70 \\
& 384 & 85.28 & 85.14 & 85.40 & 92.41 & 92.28 & 92.47 \\
& 576 & 96.02 & 95.83 & 96.09 & 98.35 & 98.12 & 98.43 \\
& 768 & 97.91 & 97.90 & 98.01 & 99.51 & 99.50 & 99.59 \\ \midrule

{\multirow{2}{*}{News}} 
& 192 & 29.52 & 29.28 & 30.14 & 37.17 & 36.51 & 37.69 \\
& 384 & 85.87 & 85.76 & 86.94 & 94.16 & 94.10 & 95.25 \\
& 576 & 96.25 & 96.18 & 97.33 & 98.01 & 98.01 & 99.10 \\
& 768 & 97.38 & 97.35 & 98.42 & 98.20 & 98.20 & 99.30 \\ \midrule

{\multirow{2}{*}{Wiki}} 
& 192 & 34.37 & 33.91 & 34.36 & 44.78 & 43.75 & 44.49 \\
& 384 & 84.71 & 84.61 & 84.76 & 92.14 & 92.00 & 92.12 \\
& 576 & 95.06 & 94.99 & 95.15 & 98.27 & 98.25 & 98.28 \\
& 768 & 98.07 & 98.02 & 98.14 & 100.00 & 100.00 & 100.00 \\ \midrule

{\multirow{2}{*}{Movies}} 
& 192 & 41.73 & 41.41 & 41.95 & 53.57 & 52.85 & 53.59 \\
& 384 & 89.45 & 89.29 & 89.76 & 96.89 & 96.84 & 97.16 \\
& 576 & 97.43 & 97.38 & 97.75 & 99.38 & 99.37 & 99.65 \\
& 768 & 98.60 & 98.59 & 98.91 & 99.57 & 99.57 & 99.84 \\ \bottomrule
\end{tabular}
\caption{Recoverability results for Phase II on LRC}
\label{tbl:LRC}
\end{table*}

\appendix
%\section{Appendices}
%\label{sec:appendix}
\section{Intrinsic Dimensionality Results}
\begin{figure}[h!]
\centering
\includegraphics[width=0.95\linewidth]{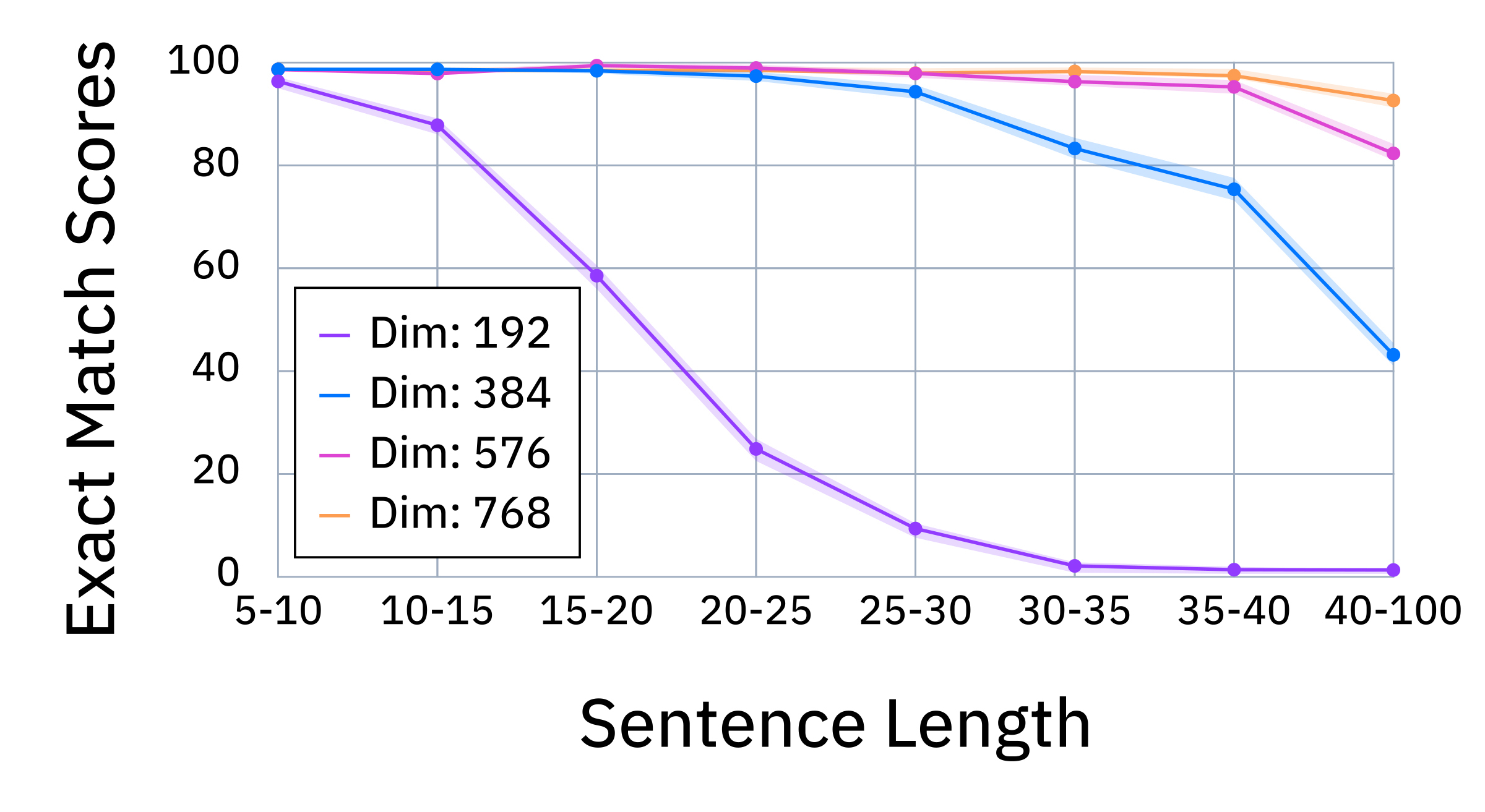}
\caption{Plot of sentence length vs. EM score on LRC for experiment IV with error regions of $\pm{\sigma}$.}
\label{em-intrinsic-dim}
\end{figure}

\begin{figure}[h!]
\includegraphics[width=0.95\linewidth]{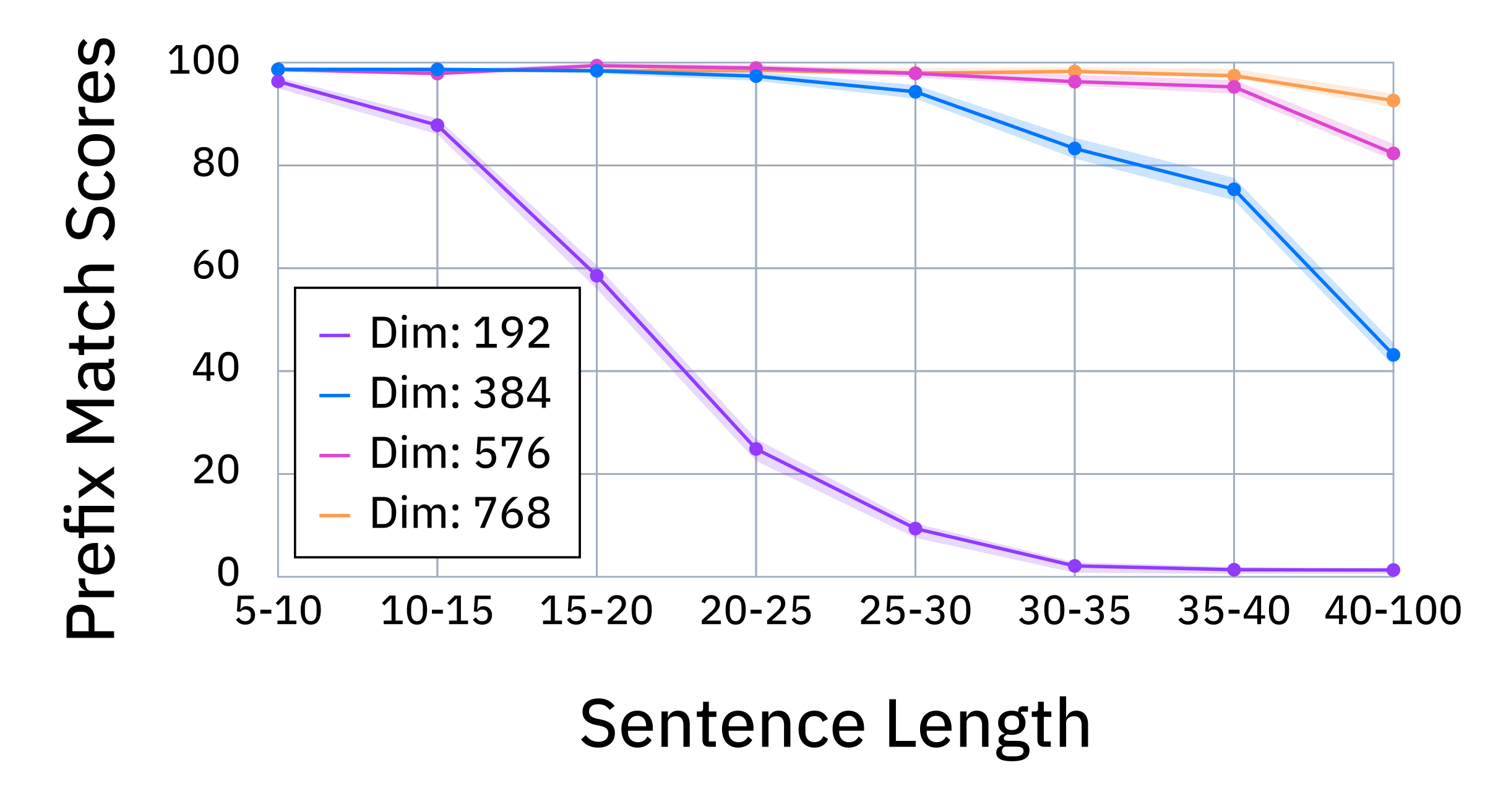}
\caption{Plot of sentence length vs. PM score on LRC for experiment IV with error regions of $\pm{\sigma}$.}
\label{pm-intrinsic-dim}
\end{figure}

We have included a table with the recoverabitility metrics for experiment IV, measuring intrinsic dimension via recoverability, from the original paper, on LRC (the large recoverability corpus).
The plot in the original paper is consistent with the results in Table~\ref{tbl:LRC}.
Recoverability performances are highest when the intrinsic dimension is close to the model's hidden dimension, $d$ (768). 
In figure~\ref{em-intrinsic-dim} and \ref{pm-intrinsic-dim} we visualize $\overline{EM}$ and $\overline{PM}$ performance scores for different intrinsic dimension $d'$ for different sentence lengths. The two plots are very similar to the $\overline{BLEU}$ vs Sentence length plot we have provided in the Results section of the paper.
Performance metrics for each corpus indicate that average recoverability over sentences is highest for the Movie dataset.
This is also consistent with $\overline{BLEU}$ by genre results we observed in the paper.


\section{Interpolation}
\begin{figure*}[h!]
\centering
\includegraphics[width=\linewidth]{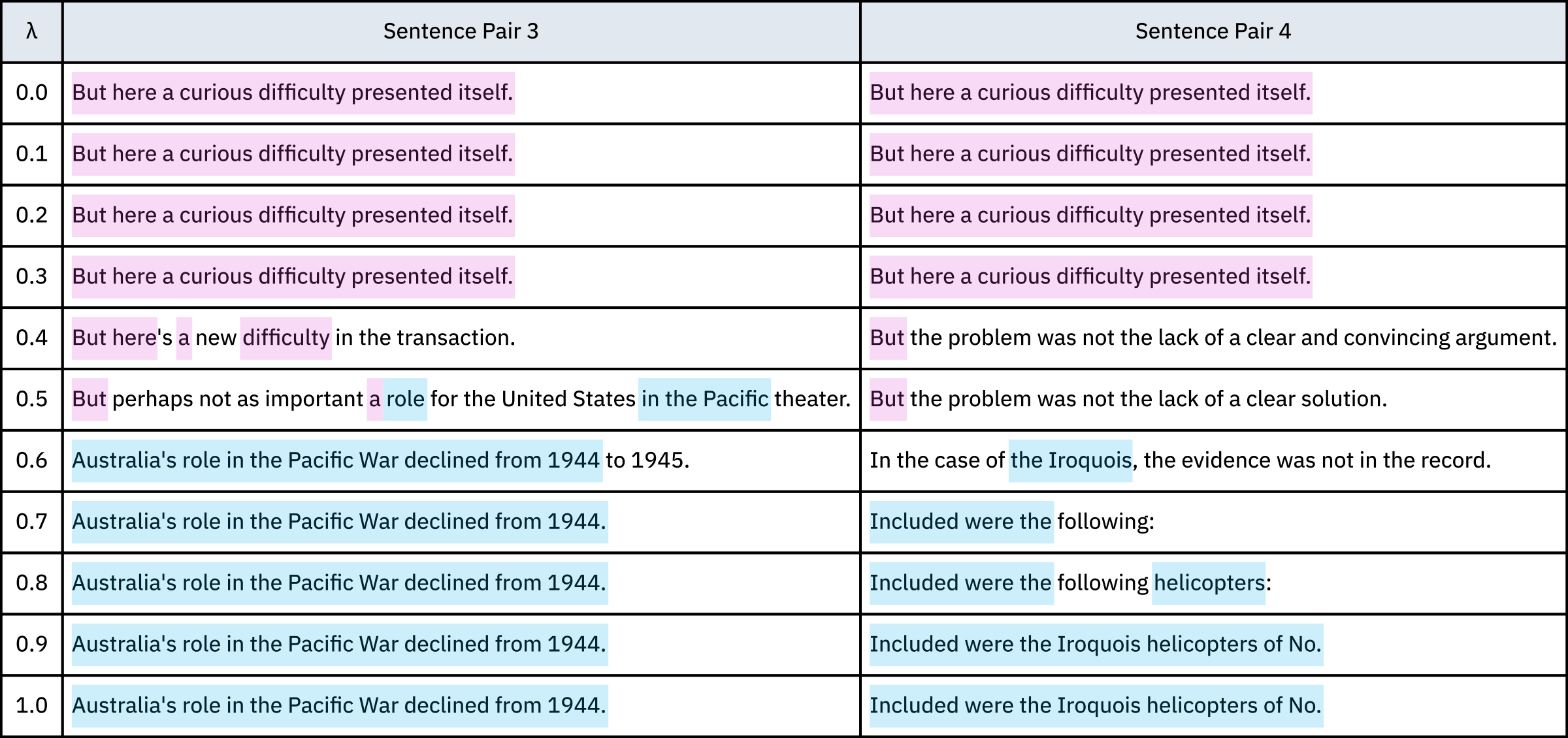}
\caption{Linear interpolations between perfectly recovered pairs of representations. Pink indicates token overlap to the first sentence, while blue indicates token overlap to the second sentence.}
\label{fig:interpolation_two}
\end{figure*}

\begin{figure*}[h!]
\centering
\includegraphics[width=\linewidth]{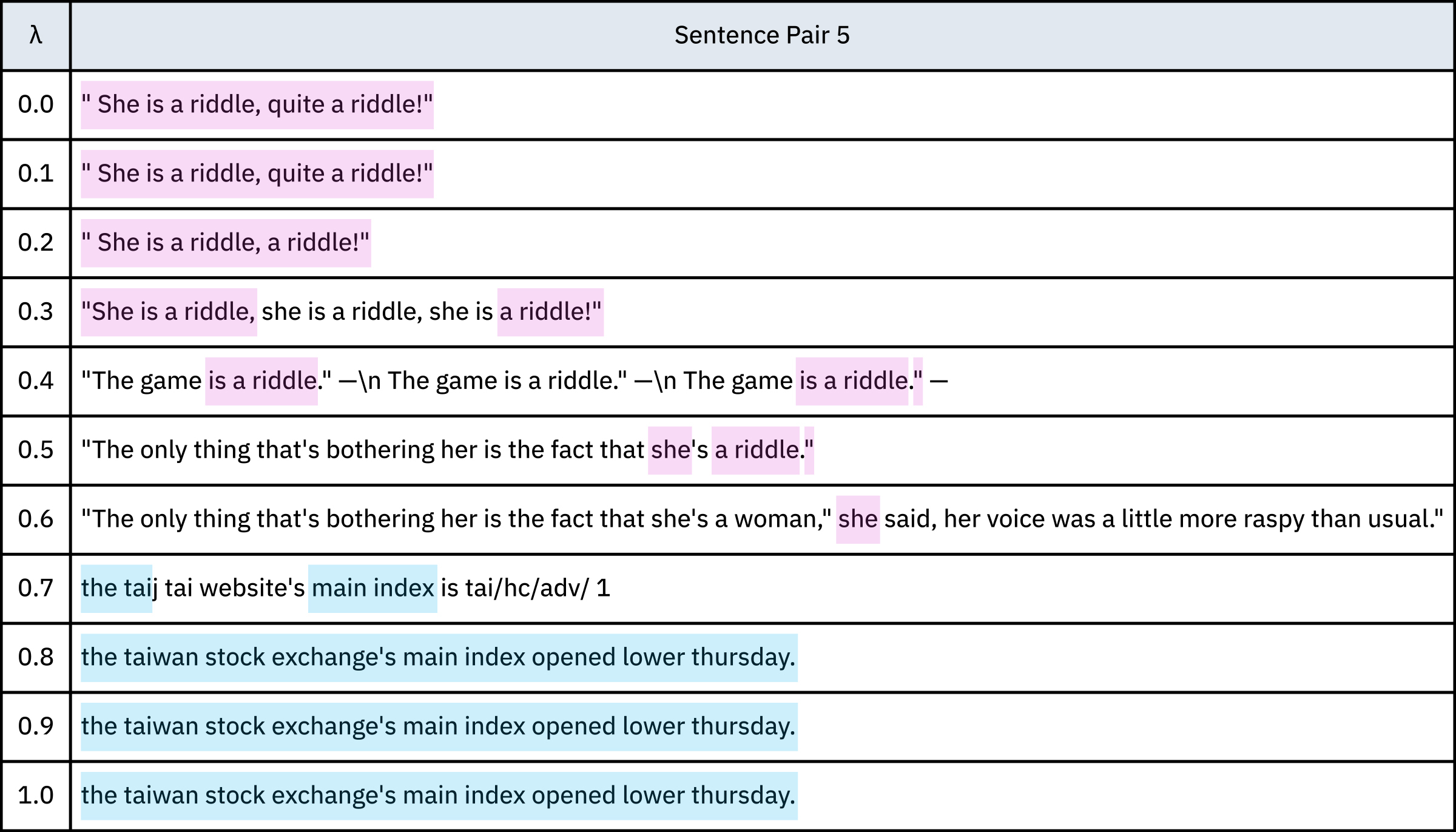}
\caption{Another linear interpolation: pink indicates token overlap to the first sentence, while blue indicates token overlap to the second sentence.}
\label{fig:interpolation_three}
\end{figure*}

\begin{figure*}[h!]
\centering
\includegraphics[width=\linewidth]{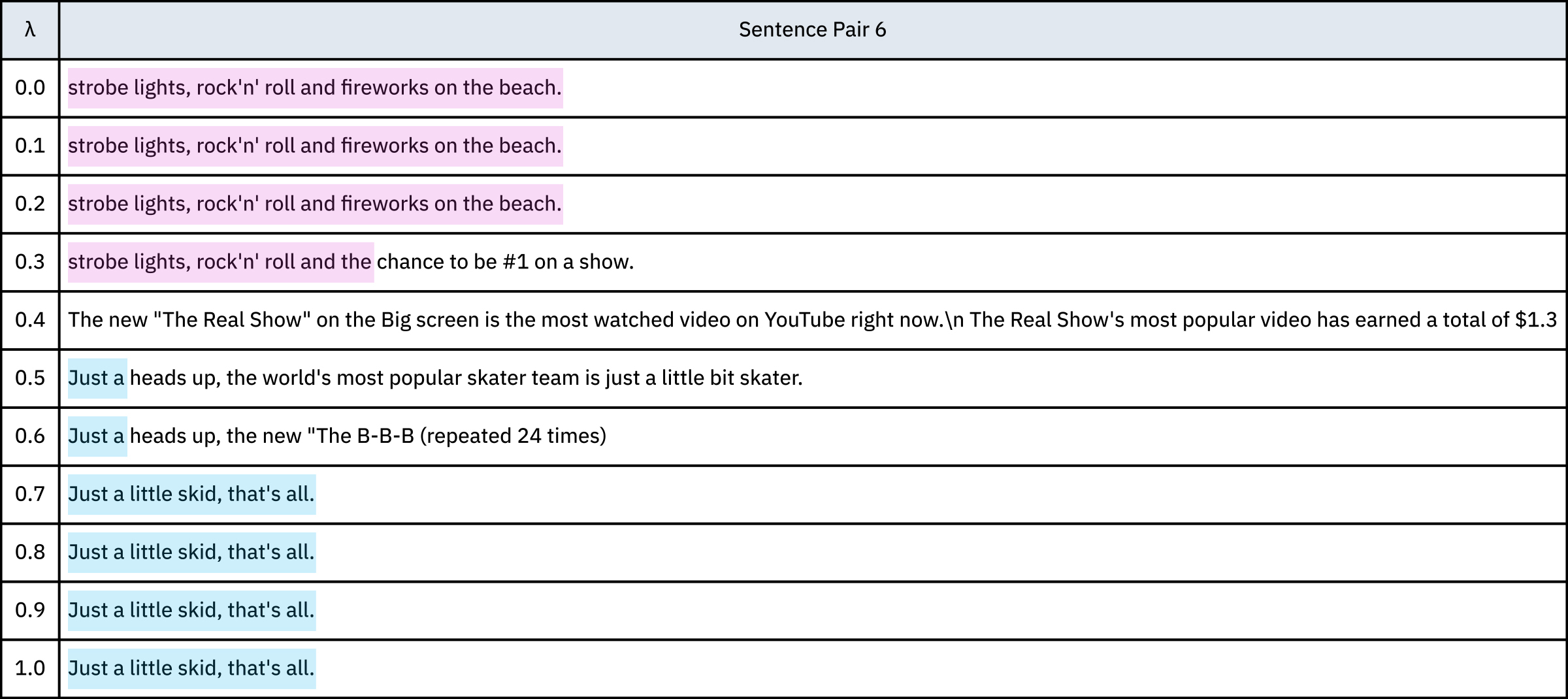}
\caption{Final linear interpolation: pink indicates token overlap to the first sentence, while blue indicates token overlap to the second sentence.}
\label{fig:interpolation_four}
\end{figure*}

We have provided some more examples of interpolation of sentence representations.
In Figure~\ref{fig:interpolation_two}, we show another two sentence pairs.
On the left, we see the same trends as we saw before with well-formed, grammatical sentences occupying every level of the interpolation.
We observe a mixing of the two sentences with lambda equaling 0.5.
One interesting finding is that the model outputs "Pacific theater," a very specific historical term used to describe World War II in the Pacific Ocean, and uses it correctly.
In the second sentence pair in Figure~\ref{fig:interpolation_two}, we observe more synonym awareness, but also observe further evidence of the nonlinearity of the sentence representation as the word "Iroquois" is forgotten when lambda equals 0.7 and 0.8.
Figure~\ref{fig:interpolation_three} shows a long sentence's representation being encoded when lambda equals 0.6 that is thematic and fluent.
Figure~\ref{fig:interpolation_four}, however, hints at the nonlinearity of the space, generating gibberish at the end with B-B-B-B repeated 24 times.

\end{document}